\documentclass[10pt,twocolumn,letterpaper]{article}

\usepackage[accsupp]{axessibility}

\usepackage[final]{cvpr}      

\usepackage{multirow}
\usepackage[dvipsnames]{xcolor}

\definecolor{cvprblue}{rgb}{0.21,0.49,0.74}
\usepackage[pagebackref,breaklinks,colorlinks,citecolor=cvprblue]{hyperref}

\def\paperID{2575} 
\def\confName{CVPR}
\def\confYear{2024}

\title{OOSTraj: Out-of-Sight Trajectory Prediction With Vision-Positioning Denoising}

\author{Haichao Zhang\\
Northeastern University\\
360 Huntington Ave Boston MA 02115\\
{\tt\small zhang.haich@northeastern.edu}
\and
Yi Xu\\
Northeastern University\\
360 Huntington Ave Boston MA 02115\\
{\tt\small xu.yi@northeastern.edu}
\and
Hongsheng Lu\\
Toyota Motor North America\\
465 N Bernardo Ave Mountain View CA 94043\\
{\tt\small hongsheng.lu@toyota.com}
\and
Takayuki Shimizu\\
Toyota Motor North America\\
465 N Bernardo Ave Mountain View CA 94043\\
{\tt\small takayuki.shimizu@toyota.com}
\and
Yun Fu\\
Northeastern University\\
360 Huntington Ave Boston MA 02115\\
{\tt\small yunfu@ece.neu.edu}
}

\begin{document}

\twocolumn[{%
\renewcommand\twocolumn [1][]{#1}%
\maketitle
\begin{center}
\centering
\includegraphics[width=0.90\linewidth]{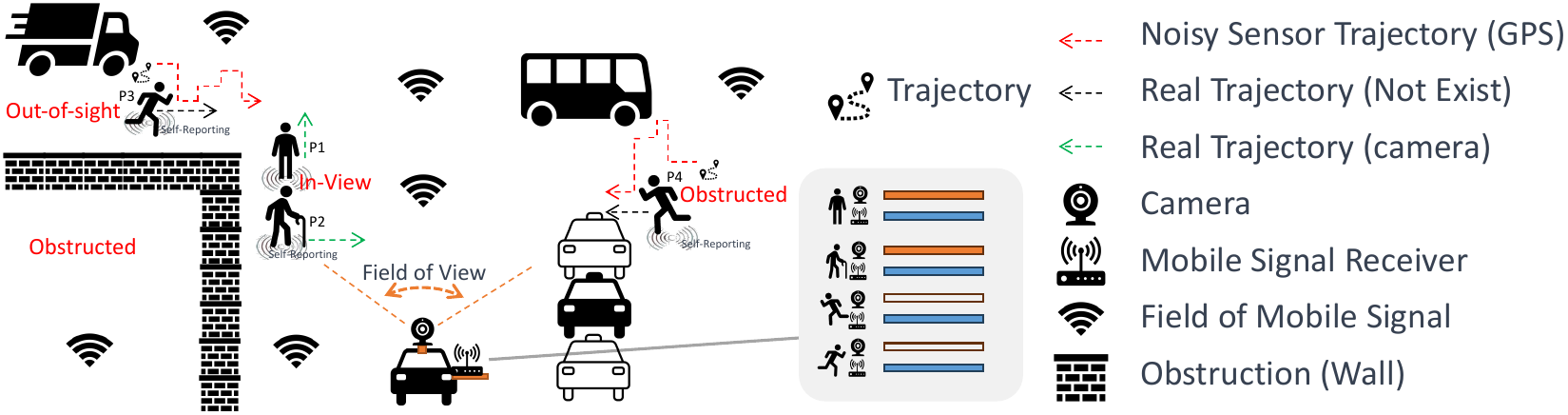}
\captionof{figure}
{
A representative illustration of real-world out-of-sight scenarios in autonomous driving. The autonomous vehicle is equipped with a camera (capturing precise visual trajectories, indicated by green dotted arrows) and a mobile signal receiver (capturing noisy sensor trajectories, represented by red dotted arrows) for tracking pedestrians and other vehicles. Pedestrians P1 and P2 are within the camera's field of view, while P3 is entirely out of sight and P4 is obscured by other vehicles. Consequently, P3 and P4 lack captured visual trajectories and are positioned dangerously, potentially crossing into the vehicle's path, posing a risk of collision. The black dotted arrows depict the hypothesized noise-free real trajectories, ideally captured by mobile sensors, contrasting with the actual noisy sensor trajectories (red arrows). The gray area in the figure demarcates the visibility range of the mobile and visual modalities: white indicates no data captured, orange signifies the presence of visual trajectories, and blue represents the availability of mobile trajectories.
}
\label{fig:head}
\end{center}%
}]




\begin{abstract}


Trajectory prediction is fundamental in computer vision and autonomous driving, particularly for understanding pedestrian behavior and enabling proactive decision-making. Existing approaches in this field often assume precise and complete observational data, neglecting the challenges associated with out-of-view objects and the noise inherent in sensor data due to limited camera range, physical obstructions, and the absence of ground truth for denoised sensor data. Such oversights are critical safety concerns, as they can result in missing essential, non-visible objects. To bridge this gap, we present a novel method for out-of-sight trajectory prediction that leverages a vision-positioning technique. Our approach denoises noisy sensor observations in an unsupervised manner and precisely maps sensor-based trajectories of out-of-sight objects into visual trajectories. This method has demonstrated state-of-the-art performance in out-of-sight noisy sensor trajectory denoising and prediction on the Vi-Fi and JRDB datasets. By enhancing trajectory prediction accuracy and addressing the challenges of out-of-sight objects, our work significantly contributes to improving the safety and reliability of autonomous driving in complex environments. Our work represents the first initiative towards Out-Of-Sight Trajectory prediction (OOSTraj), setting a new benchmark for future research. The code is available at 
\url{https://github.com/Hai-chao-Zhang/OOSTraj}.

\end{abstract}




\vspace{-3pt}
\section{Introduction}
\label{sec:intro}

Trajectory prediction is pivotal in diverse research domains, including computer vision, virtual reality, robotics, and autonomous driving. This capability is fundamental for understanding human behavior in deep learning models and crucial for enabling autonomous vehicles to anticipate pedestrian movements, thereby enhancing safety and preventing collisions. Real-world applications, however, present complex environments where trajectory prediction remains a formidable challenge. While substantial progress has been made in this field, existing approaches heavily rely on computer vision algorithms under the assumption of noise-free and constantly visible observations. This assumption often overlooks the critical aspects of out-of-sight objects and noisy data, which are prevalent in real-world scenarios. Our research addresses this gap by proposing a vision-positioning denoising method that enhances trajectory prediction in these challenging conditions.


In the context of autonomous driving and robotics, the limitations of camera-based systems pose significant challenges. As shown in Fig.~\ref{fig:head}, one common scenario involves pedestrians or vehicles being obscured by other vehicles, walls, or falling into the vehicle's blind spots. In such cases, they may suddenly appear in the camera's field of view, leaving insufficient time for deep learning algorithms to respond proactively. Additionally, the high speed of vehicles exacerbates this issue. When a camera moves rapidly, pedestrians can emerge from out-of-sight areas too quickly for the camera to capture and process their trajectory information. In both scenarios, the absence of prior visual data on pedestrians or vehicles hinders the trajectory prediction capabilities of autonomous systems. This lack of out-of-sight observations poses a serious safety risk and limits the practical deployment of autonomous driving technology in real-world situations. Our research aims to mitigate these challenges by enhancing trajectory prediction through vision-positioning denoising, enabling more reliable detection and response to such unpredictable occurrences.

Recent studies in trajectory prediction have increasingly acknowledged the impediments posed by incomplete observations, a factor that significantly hinders accurate predictions. While the bulk of existing research presumes the availability of complete and precise observational data, several noteworthy studies, such as those by Xu \etal \cite{xu2023uncovering} and Fujii \etal \cite{fujii2021two}, have begun to address this challenge. These works predominantly focus on imputing missing views or directly predicting from partial trajectory data. A notable method by Zhang \etal \cite{zhang2023layout} demonstrates the prediction of layout trajectories using just a single visual timestamp, supported by mobile sensor modalities. However, these studies share a common limitation: their dependency on a minimum quantum of visual observational data for trajectory prediction. To the best of our knowledge, there is an absence of research dedicated to predicting trajectories solely from out-of-sight observations. This research gap is not only significant in the academic field but also constitutes an urgent safety issue in autonomous driving, a concern that our research directly addresses.

Addressing the out-of-sight problem in trajectory prediction presents two key challenges. The initial challenge involves coping with noisy sensor measurements. In the absence of visual timestamps, our approach relies on commonly used localization sensors, such as GPS and odometers, to provide a basic spatial location. This reliance is necessitated by the nature of the out-of-sight problem, where traditional visual inputs are not available. However, as established in prior research, these mobile sensor modalities are subject to significant noise. GPS, for example, can exhibit errors ranging from 1 to 4 meters \cite{huang2006low}, while odometers are prone to drift noise accumulation \cite{zhao2017adaptive}. This noise significantly hampers the accuracy of sensor-derived trajectory predictions. Furthermore, the absence of ground truth for denoised sensor data complicates the situation, as it prevents us from employing supervised learning methods for denoising. Therefore, our approach involves training a denoising model in an unsupervised manner.

To mitigate this issue, we propose the use of visual information for denoising sensor data. Although our scenario lacks direct visual observations, object tracking algorithms can offer highly accurate and noise-free location data in image coordinates. By integrating this visual information, we aim to refine and denoise the trajectory data derived from the noisy sensor inputs. However, direct visual observation of out-of-sight subjects is not possible, we need to establish a connection between sensor data and visual coordinates. This brings us to our second challenge.

However, since the pedestrain is out-of-sight, to do such a denoising method with vision-positioning, another challenge is there is no visual reference even if we assume the localization is precise. To solve this challenge, we have to find a localization-vision mapping to solve the visual reference problem. Luckily, since the camera's intrinsic matrix is not changing within a frame or is not moving, we can build such localization-vision mapping by estimating the camera's intrinsic matrix. Since the pedestrians and vehicles shares the same camera's intrinsic matrix in the camera's view. To best utilize the information within camera frames, through analysis of other visible pedestrians and vehicle movements between sensor localization trajectories and visual trajectories, we can build a model to predict the mapping relationship which is highly associated with the camera's intrinsic matrix.
Nevertheless, another benefit of this design is, that if the camera is not moving in some widely used applications like video surveillance ~\cite{haering2008evolution, 10381763} or traffic monitoring~\cite{jain2019review}, we can infer the localization-vision mapping through pedestrians and vehicles in previous frames.

The contributions of this paper are as follows:
\begin{itemize}
    \item We introduce a pioneering task in the field of trajectory prediction - predicting the noise-free visual trajectory of out-of-sight objects using noisy sensor trajectories. This task (OOSTraj) addresses a critical gap in current research and extends the capabilities of trajectory prediction in complex environments. 
    \item Our work proposes an innovative vision-positioning denoising module. This module adeptly leverages precise visual data to effectively denoise sensor data with an unsupervised manner of no need for the ground truth of denoised sensor data through the camera matrix estimator, providing a much-needed visual reference for objects that are out of the camera's view. This approach marks a significant advancement in combining sensor and visual data for enhanced prediction accuracy.
    \item Through rigorous ablation studies and plug-and-play experiments, we demonstrate that our method achieves state-of-the-art results on out-of-sight noisy sensor trajectory denoising and prediction over baselines. These studies provide concrete evidence of the effectiveness and efficiency of our proposed approach.
\end{itemize}

\section{Related Works}

\subsection{Vision-Wireless Fusion}
Vision-wireless fusion represents an interdisciplinary approach that synergizes the strengths of visual and wireless (mobile) modalities to tackle inherent challenges in each. The visual modality, while offering detailed and direct observation capabilities, grapples with limitations such as obstructions, out-of-sight issues, computational intensity, and a restricted observational range. On the other hand, the wireless modality, characterized by its extensive range of sensing and communication, encounters significant issues with sensor noise~\cite{liu2016sensor} and environmental disturbances~\cite{di2002counteraction}. A persistent challenge in wireless data is the lack of ground truth, compounded by the difficulty in human labeling due to the non-visual nature of the signals.

Despite these challenges, the fusion of visual and wireless data presents a powerful tool. The visual modality, with its ease of visualization and labeling, coupled with advancements in computer vision algorithms, offers robust and precise supervision. This fusion has seen limited but impactful applications in recent studies. For instance, Liu et al.~\cite{liu2020vision} utilized wireless trajectories to address person reidentification problems, accounting for variations in posture and clothing. Alahi et al.~\cite{alahi2015rgb} (2015) employed RGB-D cameras to enhance indoor localization accuracy by estimating distances using Wi-Fi signals from smartphones. Papaioannou et al.~\cite{papaioannou2015accurate} focused on tracking people across visual and wireless modalities in dynamic industrial environments. These pioneering works have significantly advanced the field by fusing visual and mobile modalities to improve task accuracy. However, less attention has been paid to leveraging the visual modality for denoising noisy sensor trajectories, especially for out-of-sight agents, an area our work aims to explore.

\subsection{Obstructed Trajectory Prediction}
The field of obstructed trajectory prediction, while closely related, is distinct from out-of-sight trajectory prediction, which remains underexplored. Trajectory prediction has widespread applications in robotics~\cite{jetchev2009trajectory}, autonomous driving~\cite{cui2019multimodal}, and human behavior understanding~\cite{rudenko2020human}. Many existing studies~\cite{nikhil2018convolutional} operate under the assumption of complete and unobstructed observations. However, recent research has begun to acknowledge and address the challenges posed by observation obstructions. Methods such as data imputation~\cite{xu2023uncovering} and modality fusion~\cite{zhang2023layout} have been proposed to tackle these issues. Nonetheless, these approaches typically rely on having at least some observational data, leaving a gap in the context of completely out-of-sight trajectory prediction. In real-world scenarios, noisy and out-of-sight trajectories are commonplace, presenting a significant challenge for existing systems. Our work seeks to bridge this gap by employing vision positioning and denoising techniques for trajectory prediction, addressing a crucial need in the field.

\begin{figure*}[ht]
  \centering
   \includegraphics[width=0.8\linewidth]{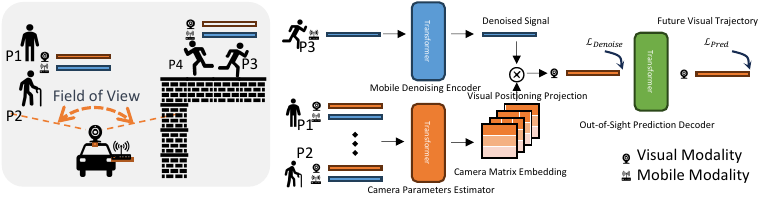}
    \caption{
    Overview of the Vision-Positioning Denoising and Predicting Model architecture. This illustration highlights the processing of pedestrian data, where pedestrians P3 and P4 are detectable only by mobile receivers, while P1 and P2 are visible to both camera and mobile receivers. The Camera Parameters Estimator Module utilizes the dual-modality trajectories of in-view pedestrians (like P1 and P2) to analyze the relationship between camera and world coordinates, resulting in a camera matrix embedding. For out-of-sight pedestrians (e.g., P3, P4), their noisy mobile trajectories are first refined by the Mobile Denoising Encoder, producing a denoised signal embedding. This embedding is then merged with the matrix embedding in the Visual Positioning Projection Module, facilitating the mapping of data into camera coordinates, with the application of $\mathcal{L}_Denoise$. Finally, the Out-of-Sight Prediction Decoder leverages the denoised visual signals to predict the trajectories of pedestrians not captured by the camera.
}
\label{fig:arch}
\end{figure*}

\section{Problem Definition}

\subsection{Symbol Annotations}
\label{Prob:Symbol}
Consider a set of \(N\) agents, represented as \(P_n, n = 1, 2, \ldots, N\), observed over timestamps \({t_1:t_2}\), starting from \(t_s\) and ending at \(t_e\). Agents visible to the camera, \(P_i\), constitute the set \(\mathbb{P}_{in}\), while those out-of-sight, \(P_o\), form the set \(\mathbb{P}_{out}\). The sensor-based noisy trajectory of the \(n\)-th agent between \(t_1\) and \(t_2\) is denoted by \(S_{P_n}^{t_1:t_2}\), which ideally corresponds to a precise but non-existent real localization trajectory, \(\hat{S}_{P_n}^{t_1:t_2}\). All agents possess a noisy sensor trajectory \(S_{P_n}^{t_s:t_e}\), significantly influenced by sensor measurement noise. Concurrently, in-view agents \(P_i, i \in \mathbb{P}_{in}\), are tracked through visual trajectories \(V_{P_i}^{t_s:t_e}\) in camera frames.

\subsection{Task Definition}
Our primary objective, given the sole input of noisy sensor trajectories \(S_{P_o}^{t_s:t_e}\), is twofold:
\begin{enumerate}
    \item Denoise the noisy sensor trajectory \(S_{P_o}^{t_s:t_e}\) without direct supervision, addressing the potential impact of sensor noise on prediction accuracy.
    \item Predict noise-free future visual trajectories \(V_{P_o}^{t_e:t_p}\) for out-of-sight agents \(P_o\) in camera frames, spanning from the observation's end timestamp \(t_e\) to the target prediction timestamp \(t_p\), using only the provided noisy sensor trajectory data.
\end{enumerate}
It is important to note that \(P_o\) lacks visual trajectories and references due to being out-of-sight. However, we can leverage other available signals, as outlined in Sec.~\ref{Prob:Symbol}, to navigate complex environmental factors.

Our strategy involves initially denoising and projecting the noisy sensor trajectory into a visual trajectory in an unsupervised manner, constructing supervision from the available signals. Subsequently, another module is employed to execute precise trajectory prediction within the visual modality.

\section{Methodology}


The overall architecture of our proposed method is detailed in Fig.~\ref{fig:arch}. Our approach is specifically designed to predict noise-free future trajectories for out-of-sight agents, utilizing noisy sensor data without relying on direct denoising supervision or visual references.

To accurately predict a noise-free trajectory from noisy sensor data, it is crucial to establish an unsupervised denoising pipeline. This pipeline is essential for filtering out noise before it can impact the trajectory prediction model. Addressing this complex challenge necessitates the integration of visual references and the construction of effective denoising supervision. We have divided this process into several distinct steps to ensure comprehensive coverage of all aspects involved in this sophisticated task.


\subsection{Mobile Denoising Encoder (MDE)}
\label{method:mde}
At the outset of our methodology, we introduce the Mobile Denoising Encoder (MDE), designed to operate under an ideal scenario where accurate supervision is available. The MDE's primary function is to denoise noisy sensor trajectories of out-of-sight agents, leveraging appropriate supervisory signals for accuracy. 

Structurally, the MDE comprises a Transformer model, flanked by two fully connected layers, both preceding and succeeding the Transformer layers. This architecture is chosen for its efficacy in handling sequential data and its capacity for capturing complex dependencies within the sensor trajectories.

In an optimal setting, the MDE processes the noisy sensor trajectories, denoted by \(S_{P_o}^{t_s:t_e}\), as its input. The objective is to refine these inputs by filtering out the noise, thereby producing what we term 'noise-free real sensor trajectories', represented by \(\hat{S}_{P_o}^{t_s:t_e}\). The operational equation of the MDE is formalized as follows:



\begin{equation}
  \hat{S}_{P_o}^{t_s:t_e}  = \mathbf{E}_{denoise} ( S_{P_o}^{t_s:t_e} ) , ~~~~~~o \sim \mathbb{P}_{out},
  \label{eq:Denoise}
\end{equation}
where the $\mathbf{E}_{denoise}$ is the MDE, the $\mathbb{P}_{out}$ is the set of out-of-sight agents, and the $P_o$ is an out-of-sight agent sampled from $\mathbb{P}_{out}$.


However, the practical implementation of the Mobile Denoising Encoder (MDE) faces a significant challenge: the ideal supervision in the form of noise-free real sensor trajectories, \(\hat{S}_{P_o}^{t_s:t_e}\), does not exist in real-world scenarios. The inherent nature of sensor measurement noise complicates the acquisition of such ideal data. Theoretically, one could envisage using higher-accuracy sensor devices to collect comparable data under identical conditions. However, this approach is impractical for most datasets and application scenarios. Even if it were feasible, these higher-accuracy sensors would likely still be subject to some level of measurement noise, albeit reduced. As a result, completely eliminating noise to construct a truly noise-free real sensor trajectory for supervision is not achievable with the current state of sensor technology.

Consequently, we are compelled to explore alternative methods for constructing effective noise-free supervision for the denoising process described in Eq. \ref{eq:Denoise}. This involves developing innovative approaches that can approximate the ideal noise-free conditions, despite the inherent limitations of existing sensor technologies.

\subsection{Visual-Positioning Denoising Module (VPD)}
\label{method:vpd}
In the Visual-Positioning Denoising Module (VPD), we leverage visual information to establish noise-free supervision for the Mobile Denoising Encoder (MDE). This approach is a response to the challenges posed by the noisy sensor modality, which is inherently prone to sensor measurement noise and difficult to refine. Visual information, on the other hand, benefits from the advancements in object tracking algorithms, offering significantly higher precision. Additionally, the nature of visual image processing, where images are discretized into integers during the rasterization process, results in visual trajectories that are highly precise, easily labelable, and effectively noise-free.

Despite the inherent noise-free quality of visual trajectories and their ideal use as supervision in trajectory prediction, a major challenge arises when dealing with out-of-sight object trajectories, which lack visual data. To address this, it is necessary to find a method to translate the denoised sensor trajectories into the visual modality effectively. This translation process is broken down into several key components: the Visual Positioning Projection Module (discussed in Sec.~\ref{method:vpp}), the Camera Parameters Estimator (outlined in Sec.~\ref{method:cpe}), and the Denoising Loss (described in Sec.~\ref{method:Dloss}). Each of these components plays a crucial role in ensuring the accurate mapping of sensor data into the visual domain, which is vital for the success of our proposed method.

\subsection{Visual Positioning Projection Module (VPP)}
\label{method:vpp}
The Visual Positioning Projection Module (VPP) plays a critical role in transforming noisy sensor trajectories into a visual format. As all visual trajectories are inherently captured by cameras, this module focuses on converting 3D world coordinates, represented by noisy sensor data, into 2D coordinates within the camera's frame. This conversion is fundamental to aligning sensor-based data with visual data, and it hinges on the principles of geometric camera calibration~\cite{strelow2001precise}.

The World-To-Camera Transformation, central to this module, is designed to compute a 2D point \(p \in \mathbb{R}^{2 \times 1}\) in the camera frame based on a corresponding 3D point \(P \in \mathbb{R}^{3 \times 1}\) in world coordinates. To facilitate this process, the height dimension in these coordinates is usually kept constant, simplifying the transformation from 3D to 2D space. The mathematical formulation of this transformation is as follows:

\begin{equation}
  [p, 1]^T = w \cdot K \cdot R_t \cdot [P, 1]^T,
  \label{eq:Calibration}
\end{equation}

where \(K \in \mathbb{R}^{3 \times 3}\) denotes the camera's intrinsic matrix, which includes parameters like the focal length and optical center that are inherent to the camera. The scale factor \(w\) adjusts the scale of the transformation to fit the camera's specific dimensions. Additionally, \(R_t \in \mathbb{R}^{3 \times 4}\) represents the extrinsic rotation and translation matrix, defining the camera's position and orientation relative to the 3D world.

To streamline this transformation process, we introduce a simplification by employing a matrix \(M \in \mathbb{R}^{3 \times 4}\). This matrix \(M\) is formulated by combining the intrinsic matrix \(K\), scale factor \(w\), and the extrinsic matrix \(R_t\). This consolidation is critical as it encapsulates all the necessary parameters and transformations required to convert a point from world coordinates into camera coordinates. The simplified form of the World-To-Camera Transformation function can be expressed as:

\begin{equation}
  M = w \cdot K \cdot R_t ,
  \label{eq:Matrix}
\end{equation}
\begin{equation}
  [p, 1]^T = M \cdot [P, 1]^T.
  \label{eq:seq}
\end{equation}

To adapt the World-To-Camera Transformation Eq.~\ref{eq:Calibration} into trajectory denoising, we need to consider the camera status in this process. If the visual trajectory is captured by a surveillance camera, the $R_t$ in $M$ will not change for engaging of timestamps, but considering the scenarios that visual trajectory is captured by a moving camera on a mobile device, we need to consider $M$ is changing for different timestamps. So, we define the sequence of $M$ as Camera Matrix Embedding $M^{t_s:t_e} \in \mathbb{R}^{(t_e-t_s)\times 3 \times 4}$ , which can be a formula as,
\begin{equation}
  M^{t_s:t_e} =  w \cdot K \cdot R_t^{t_s:t_e}    , ~~~~~~t \in [t_s, t_e],
  \label{eq:M_seq}
\end{equation}
in which $R_t^{t_s:t_e}$ and $M^{t_s:t_e}$ are extrinsic rotation \& translation matrix and Camera Matrix Embedding in timestamps between $t_s$ and $t_e$. 

We regarded noise-free real sensor trajectory $\hat{S}_{P_o}^{t_s:t_e}$ as a sequence of 3D world coordinates $P \in \mathbb{R}^{3 \times 1}$. So, combing Eq.~\ref{eq:seq} and Eq.~\ref{eq:M_seq}, the Visual Positioning Projection function to map $\hat{S}_{P_o}^{t_s:t_e}$ into visual modality can be defined as,
\begin{equation}
  V_{P_o}^{t_s:t_e} = M^{t_s:t_e} \cdot \hat{S}_{P_o}^{t_s:t_e}  , ~~~~~~t \in [t_s, t_e], o \sim \mathbb{P}_{out},
  \label{eq:vpp}
\end{equation}
in which the $V_{P_o}^{t_s:t_e}$ is the visual trajectory for out-of-sight agent $P_o$.

\subsection{Camera Parameters Estimator (CPE)}
\label{method:cpe}
Although we have the Visual Positioning Projection in Eq.~\ref{eq:vpp} to map noise-free sensor trajectory into the visual trajectory, 
the camera matrix is still not accessible because most datasets do not include the camera matrix, and it cannot be obtained from images directly. Moreover, considering the camera is moving in many practical scenarios like autonomous driving, mobile robots, and smartphones, the camera matrix may change in this process. That makes obtaining the camera matrix sequence extremely hard. 

However, since there are many in-view agents captured by the same camera in the same timestamp and have both visual and sensor trajectories, we propose a Camera Parameter Estimator to analyze the relationship between the visual and sensor trajectories. Because that relationship is highly related to the camera parameters as shown in Eq.~\ref{eq:vpp}, we predict camera matrix embedding $M^{t_s:t_e} $ from all the pairs of in-sight visual and sensor trajectories, with the Camera Parameter Estimator. The Camera Parameter Estimator consists of a transformer with a fully connected layer before and after. The prediction process can be denoted as

\begin{equation}
  M^{t_s:t_e} = \mathbf{E}_{cpe} ( V_{P_0}^{t_s:t_e}, S_{P_0}^{t_s:t_e} , ..., V_{P_i}^{t_s:t_e}, S_{P_i}^{t_s:t_e} ) ,~~~~~~ i \in \mathbb{P}_{in},
  \label{eq:cpe}
\end{equation}
where the $V_{P_i}^{t_s:t_e}$, $S_{P_i}^{t_s:t_e}$ are the visual and sensor trajectories of in-sight agents $P_i$, respectively. The ${P}_{in}$ is the set of in-sight agents. $\mathbf{E}_{cpe}$ is the Camera Parameters Estimator.

\subsection{Denoising Loss}
\label{method:Dloss}
In the last subsection, we obtain the visual trajectory of the out-of-sight agent with Eq.~\ref{eq:cpe}, since we don't have ground truth for noisy sensor trajectories and the visual modality is noise-free and easy to obtain or label as we described in Sec.~\ref{method:vpd}, we can use the process to construct a supervision for the Mobile Denoising Encoder and Camera Parameters Estimator. The denoising loss function $\mathcal{L}_{Denoise}$ can be formulated as,
\begin{equation}
  \mathcal{L}_{Denoise}  = \mathcal{L}_2 (V_{P_o}^{t_s:t_e}, \overline{V}_{P_o}^{t_s:t_e}),
  \label{eq:denoise_loss}
\end{equation}
where $ \mathcal{L}_2$ is the L2~\cite{buhlmann2003boosting} loss. The $V_{P_o}^{t_s:t_e}$ and $\overline{V}_{P_o}^{t_s:t_e}$ are the predicted visual trajectories of out-of-sight agents and the ground truth. Please note that the ground truth of the visual trajectories of out-of-sight agents is only available during training.

\subsection{Out-of-Sight Prediction Decoder (OPD)}
\label{method:opd}
We propose an Out-of-Sight Prediction Decoder to predict future visual trajectories from the predicted out-of-sight visual trajectories in Eq.~\ref{eq:cpe}, to avoid predicting directly from noisy sensor trajectories. We applied a transformer model as the prediction module for simplicity. This process can be formulated as 
\begin{equation}
  V_{P_o}^{t_s:t_p} = \mathbf{D}_{pred} (V_{P_o}^{t_e:t_p}), ~~~~~~o \sim \mathbb{P}_{out},
  \label{eq:}
\end{equation}
where the $V_{P_o}^{t_e:t_p}$ is the observation of predicted  visual trajectory of out-of-sight agent $P_o$ between timestamps $[t_s:t_e]$, and the $V_{P_o}^{t_s:t_p}$ is the predicted future visual trajectory between timestamps $[t_e:t_p]$. $\mathbf{D}_{pred}$ is the Out-of-Sight Prediction Decoder. $\mathbb{P}_{out}$ is the set of out-of-sight agents.

We applied a prediction loss $\mathcal{L}_{Pred}$ for this process,
\begin{equation}
  \mathcal{L}_{Pred}  = \mathcal{L}_2 (V_{P_o}^{t_s:t_p} , \overline{V}_{P_o}^{t_s:t_p} ),
  \label{eq:}
\end{equation}
where the $\mathcal{L}_2$ is the L2 loss. $V_{P_o}^{t_s:t_p}$ and $\overline{V}_{P_o}^{t_s:t_p}$ are the predicted and ground truth of future trajectories between $t_s$ and $t_p$ timestamp.

\subsection{Implementation Details}

The training samples of out-of-sight agents are randomly sampled from all pedestrians in a scenario, all other completed visible agents in both visual and noisy modalities are regarded as in-sight agents for the estimating of the camera parameters estimator.

The wireless data referenced in our study pertains to sensor localization signals, such as noisy GPS or odometers, typically collected by mobile phones carried by pedestrians. The vehicle, equipped with both a camera and a wireless receiver, captures sensor and visual trajectories. This data is then fed into our model for the prediction.

\section{Experiments}

\begin{table}[t]
\centering\scalebox{0.6}{
\begin{tabular}{c |c c c| c c c}
\toprule
\textbf{Dataset}    & \multicolumn{3}{c}{\textbf{Vi-Fi Dataset}~\cite{liu2022vi}} & \multicolumn{3}{c}{\textbf{JRDB Dataset}~\cite{martin2021jrdb}} \\ \toprule
\textbf{Baselines}   & \textbf{SUM$\downarrow$} & \textbf{MSE-D$\downarrow$} & \textbf{MSE-P$\downarrow$} & \textbf{SUM$\downarrow$} & \textbf{MSE-D$\downarrow$} & \textbf{MSE-P$\downarrow$}  \\ \midrule
ViTag~\cite{liu2021lost}                            & 200.90 & 100.53 & 100.37&	143.08	&	71.23	&	71.85 		\\	
Vanilla LSTM~\cite{hochreiter1997long, shi2018lstm}                     & 116.01 & 58.31  & 57.70 &	56.05	&	27.98	&	28.07		\\	
Vanilla GRU~\cite{chung2014empirical, tran2021goal}                     & 57.34  & 28.69  & 28.65 &	71.91	&	35.83	&	36.07		\\	
Vanilla RNN~\cite{rumelhart1986learning, rella2021decoder}                 & 31.61  & 15.92  & 15.69 &	112.40	&	56.00	&	56.41		\\	
Vanilla Transformer~\cite{vaswani2017attention, giuliari2021transformer}  & 28.33  & 14.26  & 14.08 &	33.37	&	16.71	&	16.66		\\	
\midrule
Ours & 27.24 & 13.42 & 13.83 &	25.51	&	10.52	&	14.99		\\	
\bottomrule
\end{tabular}}
\caption{Experiments of Quantitative Comparison of Models}
\label{tab:quan}
\end{table}


\subsection{Datasets}
\noindent\textbf{Vi-Fi Multimodal Dataset}~\cite{liu2022vi}.
The Vi-Fi dataset is a comprehensive multimodal dataset designed for vision-wireless systems, particularly focusing on linking pedestrian identities across visual and wireless modalities. It features wireless data captured from smartphones carried by pedestrians, encompassing technologies like FTM, IMU, and noisy GPS. Additionally, an RGB-D camera surveillance system, accompanied by a wireless receiver, is deployed either on a wall or on a bicycle to record pedestrian bounding boxes. This setup ensures simultaneous acquisition of both visual and wireless data. Given the nature of the data collection process, the wireless data exhibits considerable noise, a factor that cannot be overlooked.
This dataset comprises 90 sequences, each lasting approximately 3 minutes, and includes both indoor and outdoor scenarios. Indoor data were collected with five legitimate users and without any passersby, while outdoor data involved 3 actual users and 12 passersby. For our experiments, we utilized the noisy GPS signals from this dataset as representative of noisy mobile trajectories and the visual points as the visual modality. To simulate out-of-sight scenarios, we selectively obscured one pedestrian in each sequence.

\begin{table*}[t]
\centering
\scalebox{0.9}{
\begin{tabular}{c c |c c c| c c c}
\toprule
\multicolumn{2}{c}{\textbf{Dataset}}    & \multicolumn{3}{c}{\textbf{Vi-Fi Dataset}~\cite{liu2022vi}} & \multicolumn{3}{c}{\textbf{JRDB Dataset}~\cite{martin2021jrdb}} \\ \toprule
\textbf{Baselines}   &  \textbf{Add Module}   & \textbf{SUM$\downarrow$} & \textbf{MSE-D$\downarrow$} & \textbf{MSE-P$\downarrow$} & \textbf{SUM$\downarrow$} & \textbf{MSE-D$\downarrow$} & \textbf{MSE-P$\downarrow$}  \\ \midrule

\multirow{2}{*}{Vanilla LSTM~\cite{hochreiter1997long, shi2018lstm}}                    & + 2 Stage         & 118.44      & 38.85  & 79.59  &	81.22	&	39.93	&	41.28	\\	
                                                                    & + VPD (Ours)      & 30.39       & 13.76  & 16.63  &	53.16	&	11.36	&	41.80	\\	\hline
                                                                    
\multirow{2}{*}{Vanilla RNN~\cite{rumelhart1986learning, rella2021decoder}}                & + 2 Stage         & 32.16       & 17.51  & 14.65  &	138.41	&	103.97	&	34.44	\\	
                                                                    & + VPD (Ours)      & 27.78       & 13.56  & 14.22  &	31.32	&	12.31	&	19.01	\\	\hline
\multirow{2}{*}{Vanilla GRU~\cite{chung2014empirical, tran2021goal}}                    & + 2 Stage         & 40.35       & 16.88  & 23.48  &	56.56	&	17.90	&	38.66	\\	
                                                                    & + VPD (Ours)      & 28.47       & 13.69  & 14.78  &	31.70	&	11.53	&	20.16	\\	\hline
\multirow{2}{*}{Vanilla Transformer~\cite{vaswani2017attention, giuliari2021transformer}} & + 2 Stage         & 28.87       & 14.22  & 14.65  &	36.99	&	14.21	&	22.79	\\	
                                                                    & + VPD (Ours)      & 27.24       & 13.42  & 13.83  &	25.51	&	10.52	&	14.99	\\	\bottomrule 
\end{tabular}}
\caption{Experiment of Plug-and-Play on the Vi-Fi and JRDB Datasets.}
\label{tab:plugandplay}
\end{table*}

\begin{table}[t]
\centering\scalebox{0.6}{
\begin{tabular}{c |c c c| c c c}
\toprule
\textbf{Dataset}    & \multicolumn{3}{c}{\textbf{Vi-Fi Dataset}~\cite{liu2022vi}} & \multicolumn{3}{c}{\textbf{JRDB Dataset}~\cite{martin2021jrdb}} \\ \toprule
\textbf{Module Components}   & \textbf{SUM$\downarrow$} & \textbf{MSE-D$\downarrow$} & \textbf{MSE-P$\downarrow$} & \textbf{SUM$\downarrow$} & \textbf{MSE-D$\downarrow$} & \textbf{MSE-P$\downarrow$}  \\ \midrule
    w/o CPE in Sec.~\ref{method:cpe}                & 32.12 & 17.01 & 15.12 &	29.66	&	14.35	&	15.31	\\			 	 	 
    w/o MDE in Sec.~\ref{method:mde}                & 32.33 & 17.65 & 14.68 &	32.94	&	17.90 	&	15.03	\\	
    w/o VPP in Sec.~\ref{method:vpp}                & 28.42 & 14.05 & 14.37 &	32.48	&	14.45	&	18.03	\\
    w/o OPD in Sec.~\ref{method:opd}                & 27.33 & 13.65 & 13.68 &	30.01	&	14.99	&	15.02	\\	
    \midrule
    Full Model & 27.24 & 13.42 & 13.82 &	25.51	&	10.52	&	14.99	\\	
\bottomrule
\end{tabular}}
\caption{Ablation Study of Module Components.}
\label{tab:ablation}
\end{table}

\noindent\textbf{JackRabbot Dataset (JRDB)}~\cite{martin2021jrdb}.
The JRDB dataset, collected by the social mobile robot JackRabbot, is designed for autonomous robot navigation and social robotics studies within human environments. It features 60,000 annotated frames, including 2.4 million 2D and 2.8 million 3D human bounding box annotations, captured using 360 degree RGB (5 cameras) and LIDAR point cloud technology. This dataset encompasses both obstructed and visible pedestrians, with the potential for each camera to have blind spots relative to the others. The bounding boxes are manually labeled, albeit at a lower annotation rate, and are subject to noise due to upsampling via linear interpolation. Size and location estimates of obstructed pedestrians in point clouds are manually conducted. For our experiments, we use the center points of 3D bounding boxes as noisy sensor trajectories and visual points as visual trajectories. To simulate out-of-sight scenarios, we randomly obscured a pedestrian, using the trajectories of several other visible pedestrians to estimate camera parameters in our model.

\subsection{Evaluation Setup}
In both the JRDB and Vi-Fi datasets, we observe 100 timestamps of out-of-sight trajectories and predict the subsequent 100 timestamps in the visual modality. Simultaneously, we input 100 timestamp pairs of in-sight visual trajectories and noisy sensor signals into our camera parameters estimator to obtain the camera matrix embeddings.

\subsection{Metrics} 
As the first study to denoise out-of-sight sensor trajectories into a visual modality for noise-free trajectory prediction, we adapt the Mean Square Error per Timestamp (MSE-T) metric from ~\cite{zhang2023layout}. This metric calculates the average pixel distance over the timestamp dimension. Given our task's dual focus on denoising and prediction performance, we apply MSE-T to both projected out-of-sight visual trajectories and predicted visual trajectories, denoting these as \textbf{MSE-D} and \textbf{MSE-P}, respectively. Additionally, we introduce the \textbf{SUM} metric, which aggregates MSE-D and MSE-P, to provide an overall score balancing denoising and prediction efficacy.

\subsection{Baselines}
To benchmark our method in noisy sensor trajectory denoising and out-of-sight trajectory prediction, we select several baselines, encompassing fundamental models like Vanilla LSTM~\cite{shi2018lstm}, Vanilla RNN~\cite{rella2021decoder}, Vanilla GRU~\cite{tran2021goal}, and Vanilla Transformer~\cite{giuliari2021transformer}. We also include ViTag~\cite{liu2021lost}, the state-of-the-art method on the Vi-Fi dataset, modified to align with our task outputs.

\subsection{Quantitative Comparison Experiments}
We conducted quantitative experiments to compare our method's performance against five baselines across two datasets and three metrics, as detailed in Table.~\ref{tab:quan}. Our model outperforms the baselines, showcasing its superiority. Notably, although ViTag is recognized as the state-of-the-art on the Vi-Fi dataset for visual and wireless identity association, it underperforms in our task. A key finding is that our model demonstrates more pronounced superiority in the denoising task than in the prediction task, underscoring the complexity and difficulty in denoising noisy sensor trajectories and mapping out-of-sight objects into the visual modality.
Furthermore, a positive correlation between denoising and prediction performance is observed among the baselines. Improved denoising leads to enhanced prediction accuracy, validating our approach of focusing on sensor data denoising to facilitate better out-of-sight visual trajectory prediction.

We also note varying performances of the baselines across different datasets. For instance, Vanilla RNN excels in the Vi-Fi dataset but falls short in the JRDB dataset, likely due to differing sensor noise characteristics in these datasets. Our vision-positioning denoising model, tailored to handle various sensor noises with precise visual modality supervision, consistently achieves robust denoising performance across different datasets and scenarios. This hypothesis is further explored in our Plug-and-Play experiment.

\subsection{Plug and Play Experiments}
In the Plug-and-Play experiment, outlined in Table.~\ref{tab:plugandplay}, we assess whether our vision-positioning denoising model enhances the performance of baselines in both noisy sensor trajectory denoising and future visual trajectory prediction. Baselines were modified into a 2-stage format for a comparative analysis. The first stage focuses on denoising, followed by prediction in the second stage. The "+ VPD" row features our vision-positioning denoising (VPD) model, as described in Sec.~\ref{method:vpd}, concatenated with the baseline to evaluate the added improvement.

Results in Table.~\ref{tab:plugandplay} indicate that incorporating our VPD model significantly enhances baseline performance in both denoising and prediction tasks. Notably, Vanilla RNN, which struggles with JRDB's sensor noise, shows marked improvement in the Vi-Fi dataset when paired with our VPD. This improvement in denoising subsequently boosts prediction accuracy. These findings affirm the efficacy of our model in augmenting denoising and prediction capabilities of existing models in a plug-and-play approach.


\subsection{Ablation Study}
An ablation study was conducted to evaluate the contribution of each module component. This was achieved by systematically removing each component, with the results presented in Table.~\ref{tab:ablation}. The study reveals that each component plays a vital role in the overall performance of the model.

The absence of the Camera Parameter Estimator (CPE) leads to a significant performance drop. Without the CPE, the model struggles to establish the visual-positioning relationship through camera matrix estimation, reducing it to a mere two-stage encoder-decoder architecture akin to the baselines. Similarly, the removal of the Mobile Denoising Encoder (MDE) results in an even more pronounced decline in performance. Without the MDE, the model faces challenges in denoising the mobile noisy trajectories, inadvertently allowing noise to propagate into the Visual Positioning Projection (VPP) process.
The VPP module, responsible for projecting out-of-sight sensor trajectories into visual trajectories, proves to be crucial. The notable performance drop in scenarios without VPP underscores the importance of explicitly incorporating knowledge from geometric camera calibration into the model. Additionally, the absence of an out-of-sight prediction decoder slightly reduces the model's effectiveness, as it then needs to infer the entire sequence during the denoising process.
Overall, this ablation study clearly demonstrates the necessity and significance of each component within our model for achieving noise-free out-of-sight trajectory prediction.

\section{Conclusion}

Our study tackles a significant challenge in trajectory prediction, specifically in situations where agents are entirely out of sight, making visual observations unfeasible. We introduce an innovative task that focuses on predicting the visual trajectories of such out-of-sight agents by utilizing noisy sensor data. The cornerstone of this task is our Visual-Positioning Denoising Module, which adeptly addresses the absence of visual references. This is achieved through sophisticated camera calibration techniques, enabling us to estimate camera matrix sequences and establish a vision-positioning mapping. This method effectively denoises sensor trajectories without relying on direct supervision, instead utilizing supervision constructed from the visual positioning projection.

Our experimental results robustly demonstrate the effectiveness of our proposed pipeline and architecture, significantly outperforming existing methods in predicting trajectories of agents that are not visible. The inclusion of visual positioning in the denoising process markedly improves the accuracy of our trajectory predictions. To the best of our knowledge, this is the first study to successfully predict out-of-sight trajectories using noisy sensor data and to employ vision-positioning projection for the purpose of denoising these trajectories. The methodologies and insights gleaned from our work pave the way for further research and development in the trajectory prediction field, especially in real-world scenarios where complete visibility of agents is not guaranteed, or where sensor data are inherently noisy and challenging to denoise.

\section{Acknowledgements}
We gratefully acknowledge the support and sponsorship provided by Toyota Motor North America for this research.

{
    \small
    \bibliographystyle{ieeenat_fullname}
    \bibliography{main}
}


\end{document}